\documentclass[runningheads]{llncs}
\usepackage{graphicx}

\usepackage{tikz}
\usepackage{comment}
\usepackage{amsmath,amssymb} %
\usepackage{color}

\usepackage[accsupp]{axessibility}  %

\usepackage{enumitem}
\usepackage{booktabs}
\usepackage[pagebackref,breaklinks,colorlinks,hypertexnames=false]{hyperref}

\usepackage{caption}
\usepackage{subcaption}
\usepackage{multirow}
\usepackage{pifont}%
\newcommand{\xmark}{\ding{55}}%

\usepackage[capitalize]{cleveref}
\crefname{assumption}{Assumption}{Assumptions}

\newcommand{\firstmyparagraph}[1]{\par\noindent \textbf{#1}}
\newcommand{\myparagraph}[1]{\vspace{0.75ex}\par\noindent \textbf{#1}}
\DeclareMathOperator*{\argmax}{arg\,max}
\DeclareMathOperator*{\argmin}{arg\,min}

\RequirePackage{xspace}
\makeatletter
\DeclareRobustCommand\onedot{\futurelet\@let@token\@onedot}
\def\@onedot{\ifx\@let@token.\else.\null\fi\xspace}

\def\eg{\emph{e.g}\onedot} 
\def\ie{\emph{i.e}\onedot}

\makeatother

\begin{document}
\pagestyle{headings}
\mainmatter
\def\ECCVSubNumber{6956}  %

\title{Totems: Physical Objects for Verifying Visual Integrity} %

\titlerunning{Totems: Physical Objects for Verifying Visual Integrity}
\author{Jingwei Ma\inst{1} \and
Lucy Chai\inst{2} \and
Minyoung Huh \inst{2} \and
Tongzhou Wang \inst{2} \and
Ser-Nam Lim\inst{3} \and
Phillip Isola \inst{2} \and
Antonio Torralba \inst{2}
}
\authorrunning{J. Ma et al.}
\institute{\textsuperscript{1}University of Washington \enspace \enspace \enspace \textsuperscript{2}MIT \enspace \enspace \enspace \textsuperscript{3}Meta AI}
\maketitle

\begin{center}
    \centering
    \includegraphics[width=1.0\textwidth]{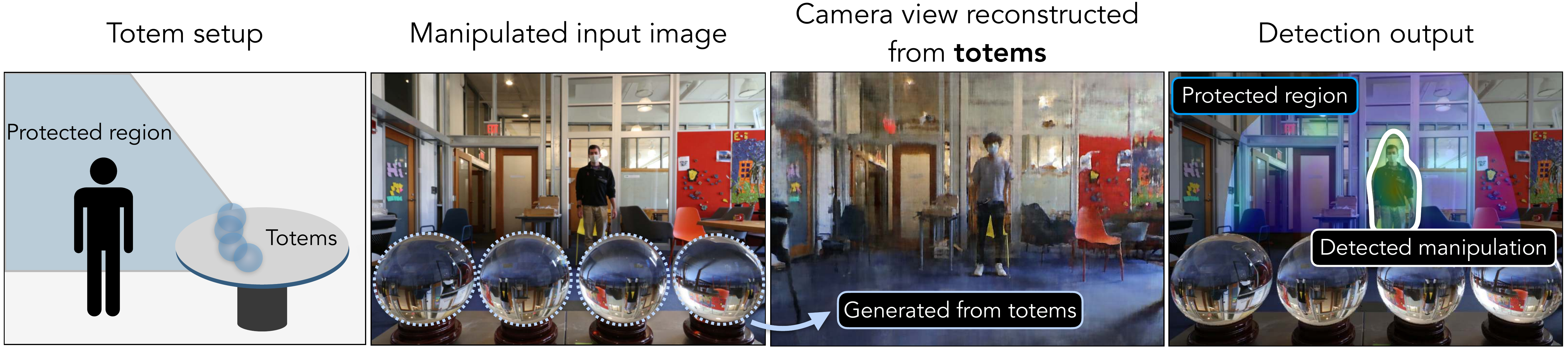} %
    \captionof{figure}{\small We envision a setup in which physical objects, called totems, are placed into a scene to protect against adversarial manipulations. From a single camera capture, the totems provide alternative distorted views of the image, which allows us to reconstruct the underlying 3D scene. The reconstruction is then used to highlight potential image manipulations by comparing the scene viewed from the totems to the image observed by the camera.
    \label{fig:teaser}}
\end{center}%

\begin{abstract}
We introduce a new approach to image forensics: placing physical refractive objects, which we call totems, into a scene so as to protect any photograph taken of that scene. Totems bend and redirect light rays, thus providing multiple, albeit distorted, views of the scene within a single image. A defender can use these distorted totem pixels to detect if an image has been manipulated. 
Our approach unscrambles the light rays passing through the totems by estimating their positions in the scene and using their known geometric and material properties. To verify a totem-protected image, we detect inconsistencies between the scene reconstructed from totem viewpoints and the scene's appearance from the camera viewpoint. Such an approach makes the adversarial manipulation task more difficult, as the adversary must modify both the totem and image pixels in a geometrically consistent manner without knowing the physical properties of the totem. Unlike prior learning-based approaches, our method does not require training on datasets of specific manipulations, and instead uses physical properties of the scene and camera to solve the forensics problem.

\end{abstract}

\section{Introduction}
\label{sec:intro}

As new technologies for photo manipulation become readily accessible, it is vital to maintain our ability to tell apart real images from fake ones. Yet, the realm of current image verification methods is mostly \textit{passive}. From the point of view of a person who wants to be protected from adversarial manipulation, they must trust that the downstream algorithms can recognize the subtle cues and artifacts left behind by the manipulation process. 

How can we give a person \textit{active} control over maintaining image integrity? Imagine if one could place a ``signature'' in a scene before the photograph is taken. %
Then, the verification process simply becomes the task of checking whether the signature matches the image content. Inspired by the movie \textit{Inception}, where characters use unique \textit{totems} to distinguish between the real world and the fabricated world, we propose to use physical objects as totems that determine the authenticity of the scene. %
In our setting, a \textit{totem} is a 
refractive object %
that, when placed in a scene, displays a distorted version of the scene on its surface (a \textit{totem view}). After a photo is taken, the defender can check whether the scene captured by the camera is consistent with the totem's appearance.

\begin{figure*}[t!] %
\centering
\includegraphics[width=1\textwidth]
{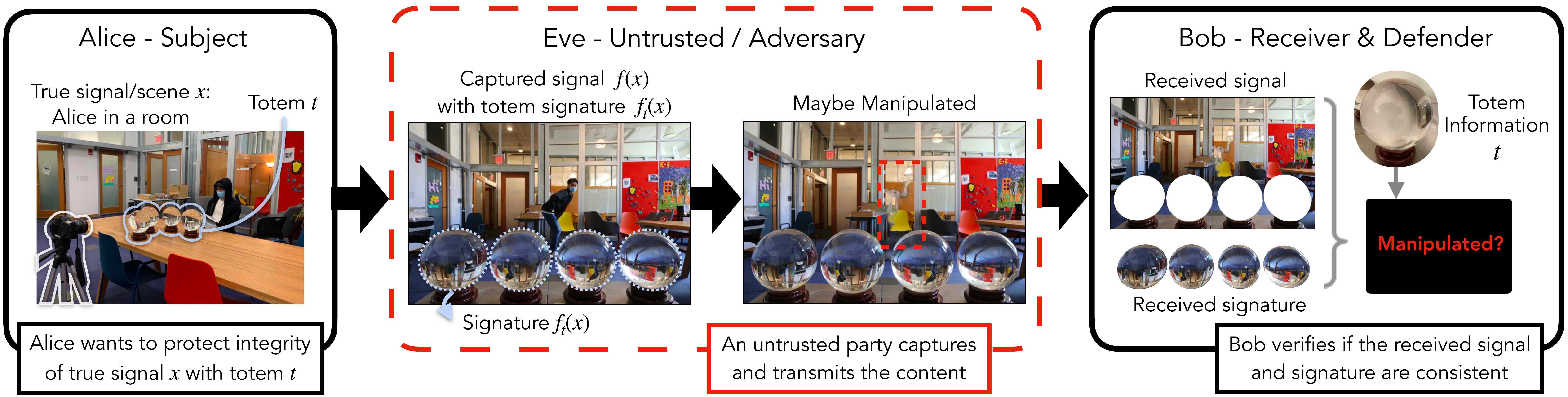}\vspace{-0.25cm}
\caption{\small Totems as a general method to protect a piece of signal (of Alice). Even if the middle person (Eve) can tamper the content, they can't easily edit both the signal and the totem signature consistently. A defender (Bob) can thus detect such manipulations. }\label{fig:alice-eve-bob}\vspace{-3pt}
\end{figure*}

However, decoding the totems and verifying consistency from a single image presents several challenges. The image of the scene observed through the totem depends on: the totem's physical properties (\eg, index of refraction (IoR)), the totem's position from the camera, and the scene geometry. We assume that the defender \emph{exclusively} knows the physical properties of the totem as the ``key'' needed to unscramble the totem views. 
The critical assumption here is that, from the adversary's perspective, it is extremely difficult to manipulate the image and the totem in a geometrically consistent manner without having access to the totem's properties. For the defender, knowing the totem's physical properties makes it possible to estimate the totem's position within the scene and to check for consistency between the scene and the totems.
In fact, this process can be bolstered by further complicating the adversary's job, using either multiple totems or a single totem with complicated facets that act to ``encrypt" the scene.

More generally, the totems can be viewed as signatures that protect the identity of a piece of signal/information, as shown in \Cref{fig:alice-eve-bob}.
In cryptography, digital signatures also act as a form of active defense, where a message is always sent together with a signature, used to verify both message integrity and sender identity (\eg, the signature can be private-key encoding) \cite{diffie1976new}. The totems are conceptually similar to such signatures, but are also fundamentally different in that (1) they give control to the subject rather than the party who captures/transmits the photo and (2) they are physical objects and fill the ``analog hole'' \cite{haber2003if}, the phenomenon that digital protections become invalid once the content is converted to analog form (\eg, via printing).

In this work, we explore an initial realization of a totem-based verification system where we use simple spherical totems. %
Our contributions can be summarized as follows:
\begin{itemize}[topsep=-2.1pt, itemsep=-1pt, parsep=0.5ex]
    \item We propose an \textit{active} image verification pipeline, in which we can place physical objects called totems within a scene to certify image integrity. 
    \item Totems create multiple, distorted projections of a scene within a single photo. Without access to totem physical properties, it becomes difficult to manipulate the scene in a way that is consistent across all totems.
    \item For verification, we undo the distortion process and infer the scene geometry from \textit{sparse} totem views and \textit{unknown} totem poses by jointly learning to reconstruct the totem pixels and optimizing the totem pose. 
    \item Comparing the scene reconstructed from the totems and the pixels of the camera viewpoint enables us to detect manipulation from a single image.
    \item Our work is an initial step towards hardware and geometry-driven approaches to detecting manipulations. The proposed framework is not specific to the scene reconstruction method we test in this paper, and may become more robust as more advanced scene reconstruction methods become available.
\end{itemize}

\section{Related Work}
\firstmyparagraph{Learning properties of refractive objects.} 
Compared to opaque materials, refractive and transparent objects pose a unique set of challenges in vision and graphics due to their complex interactions with light rays. For example, reconstructing the shape of refractive objects requires multiple views of the refractive object from different viewing angles and makes several assumptions about the environment and capture setup, such as a moving camera and parametric shape formulation~\cite{BenEzra2003WhatDM}, known IoR and correspondences between scene 3D points and image pixels~\cite{kutulakos}, or known object maps, environment map, and IoR~\cite{li2020through}. Another line of work models the light transport function in refractive objects. These approaches also involve placing planar backgrounds behind the object and capturing multiple viewpoints, which then allow for environment matting to compose a refractive object with different backgrounds~\cite{Wexler2002ImagebasedEM,envmat}. In our work, we assume that the knowledge about the totem is asymmetrical between the defender and adversary. The defender knows full totem specifications such as IoR, shape, and size, but the adversary must guess these parameters. Thus, the defender is better able to model the light refraction process through the totems compared to the adversary. Further, we do not require a specific capture environment beyond a single image taken with multiple totems visible in the scene.

\myparagraph{Accidental cameras.} Oftentimes, the objects around us can form subtle, unexpected cameras. However, decoding the image from these non-traditional cameras is much more challenging than reading directly from a camera sensor. We are inspired by the classic work of Torralba and Freeman~\cite{torralba2012accidental}, which uses shadows within a room to recover a view of the world outside. By observing changes in the indirect illumination within a room, it is then possible to infer properties such as the motion of people outside of the camera frame~\cite{aittala2019computational,sharma2021you}. Using specular objects as distorted mirrors, Park et al.~\cite{park2020seeing} recover the environment by looking at an RGB-D sequence of a shiny input object, while Zhang et al.~\cite{zhang2015sparklevision} decode images from the reflection pattern of randomly oriented small reflective planes comprised from glitter particles. Information about scenes is also inadvertently contained in sparse image descriptors, such as those from a Structure-from-Motion (SfM) point cloud, and can be used to render the scene from a novel viewpoint~\cite{pittaluga2019revealing}. With respect to transparent or semi-transparent objects, decoding textures such as water droplets on a glass surface can reveal the structure of the room behind it~\cite{IseringhausenSIG2017}, even if the glass is intentionally obscured using that texture~\cite{10.1007/978-3-642-15567-3_27}. These hidden cameras have serious implications towards privacy, but here we leverage totems as a hidden camera for an alternative purpose -- verifying the integrity of possibly manipulated images using a multiview consistency check.

\myparagraph{Detecting image manipulations.} There are numerous ways to edit an image from its original state, warranting a large collection of works that identify artifacts left behind by various manipulation strategies. A number of manipulation pipelines involve modifying only part of an image (e.g. cut-and-paste operations and facial identity manipulations~\cite{rossler2019faceforensics++,dolhansky2019deepfake}), and thus detection approaches can either directly identify the blending~\cite{li2020face} or warping artifacts~\cite{wang2019detecting,li2018exposing} or leverage consistency checks between different parts of the image to locate the modified region~\cite{huh2018fighting,zhao2021learning,zhou2017two,mayer2020exposing,fu2012,Wu2019ManTraNet}. Our approach intends to build the consistency check into captured image, rather than using a learned pipeline. Other cues for detection include subtle traces left behind by the camera or postprocessing procedures~\cite{agarwal2017photo,johnson2006exposing,popescu2005exposing},
or human biometric signals~\cite{agarwal2019protecting,li2018ictu,yang2019exposing}. %
With the rise of image synthesis techniques and image manipulation using deep neural networks, it has been shown that the architectures of these networks also leave detectable traces~\cite{wang2020cnngenerated,zhang2019detecting,yu2019attributing,marra2019gans}, and that image generators can also reflect signatures embedded in the training data~\cite{yu2021artificial}. Another way to verify image integrity is to assume that we have multiple viewpoints of the same scene, captured at the same time; then detecting inconsistencies among these different viewpoints can signal potential manipulations~\cite{tursman2020towards,zhang2010}. Our setup is most similar to these latter approaches, but we relax the assumption of having multiple cameras and instead use refractive totems to obtain multiple projections of the scene within a single image. Moreover, using irregular totems as distorted lenses may further increase the difficulty of successful manipulations. %

\myparagraph{Digitial signatures, cryptography, and physical one-way functions.} Our general Totem framework (see \Cref{sec:general-totem}) is conceptually similar to digital signature schemes in cryptography \cite{diffie1976new}. Both ideas add a certificate/signature to a message, which is then used by the recipient to verify message integrity. However, as mentioned in the introduction, totems are physical objects that give control to the subject (rather than photographer/transmitter) and fill the ``analog hole'' \cite{haber2003if}. Additionally, unlike digital signatures, they are not restricted to a particular sender and thus does not verify sender identity. This paper focuses particularly on an instantiation of this general framework in the visual domain, where views via distorted lenses are assumed to be hard to manipulate. Also utilizing physical behaviors of complex material, Pappu~et~al.~\cite{pappu2002physical} showed promising results in creating physical processes with cryptographical properties, termed physical one-way functions, for their easiness to evaluate and hardness to invert. While such processes are complex and not readily suited for image verification, they may have potential implications in future totem geometry designs.

\section{The Totem Verification Framework}

\subsection{Physical Totems for Image Verification}

In image manipulation, the adversary modifies the content of a single image with the intent that a viewer would infer a different scene than the one originally depicted by the unmodified image. For instance, the perpetrator of a crime might edit themselves out of a photo of the crime scene. %

Easy access to modern image-processing software and deep image manipulations has significantly lowered the barrier to making such realistic attacks on a single image \cite{photoshop,dolhansky2019deepfake}. Is there really no hope to defend against such attacks and protect the integrity of photographs we have taken? In this framework, we propose a potential solution. 

We argue that a single camera image is a rather vulnerable format as it only represents \emph{one} view of the underlying scene that we want to verify. 
Moreover, the camera's mapping from scene to the image is well-understood, so it is rather easy to infer the scene and edit the image. 
But what if the defender also receives other views of the same scene, where the lenses are customized such that the mappings from scene to such views are only known to the defender? Indeed, it would be harder for an adversary to provide such a set of views that \emph{consistently} represent a different scene.

After all, simply obtaining multiple photos of the same scene is no easy feat, often requiring coordination among multiple cameras, let alone requiring custom lenses. In comparison, most current image-hosting services only ask for a single-view image, which can be captured by everyday devices such as cellphones. Therefore, we desire a set-up where:
\begin{itemize}[topsep=-2.1pt, itemsep=-1pt, partopsep=9pt,parsep=0.5ex]
    \item The content includes various distorted views of the scene from different spatial locations and angles, 
    \item The distortions (\ie, lens properties) are only known to the defender,
    \item The process does not require significant equipment investment, high skill, or an obscure content format.
\end{itemize}\vspace{3pt}
Essentially, such a mechanism can be accessible to common users who create and upload visual content without adding much complications to their workflow.

Towards these goals, we propose placing small refractive objects in the scene and capturing the image as usual. The appearance of these objects in the camera image essentially forms small lenses of the same scene from different locations and angles (see~\Cref{fig:teaser}). Such objects can be custom designed and mass-produced with simple materials (\eg, glass) to be widely and cheaply available. Therefore, with the same imaging devices and file format, the uploaded image now itself contains multiple (distorted) views of the same scene. Unlike traditional digital signatures in which the photographer generates a certificate as a post hoc procedure, we use a single certificate (\eg, totems) -- often owned by the subject or defender -- that is used across all photos, giving an active control to the subject that is getting photographed.

With exclusive knowledge of these objects' physical properties, the defender may then extract such multiple views and check if they, and the rest of the normal camera view, can form a consistent 3D scene. 
The physical laws and specific properties of these objects place various constraints on these views. 
The defender may check specific constraints, or attempt a multiview 3D scene reconstruction, as we explore in \Cref{sec:method}. 
Notably, such procedures are not available to adversaries who do not have access to the detailed object properties.

\subsection{The General Totem Framework}\label{sec:general-totem}

The above image verification procedure is an instance of a more general approach  (\Cref{fig:alice-eve-bob}). There, we assumed that physical rendering through custom ``lenses'' is a process that is hard to manipulate but easy to verify. In general, similar processes can be used for verifying the integrity of various data modalities.

\myparagraph{Setting.} A true signal $x$ (\eg, a 3D scene) is conveyed via a compressed format $y = f(x)$ (\eg, a 2D image). An adversary may manipulate $y \rightarrow y'$ such that receiver of $y'$ believes that it represents some different signal $x'$ with $y' = f(x')$ (\eg, editing an image to depict a different scene).

\myparagraph{Defense.} To detect such attacks, a defender provides a \emph{totem} $t$ and requires every submission of $y=f(x)$ to be accompanied with $f_t(x)$ as a certificate of $x$ generated with $t$ (\eg, totem views of the scene). Now, an adversarial attack will have to manipulate $(f(x), f_t(x))$ to $(f(x'), f_t(x'))$ (without having $x'$). This would be a much harder task if $f_t$ is sufficiently complex, because the adversary now have to forge a consistent pair $(f(x'), f_t(x'))$, without knowledge of $f_t$'s internal logic. On the other hand, the defender, with access to detailed knowledge of $t$ and $f_t$, can verify if they have received a consistent pair. The extended submission $(f(x), f_t(x))$ ideally should be easy to create (with $x$ and $t$) and not require much more overhead (than just $f(x)$).

The specific totems (and corresponding method of verification) depend on the type of signal. In the present paper, we focus on visual signals, where we can use the physical laws of rendering / imaging. Similarly, distorted audio reflectors could be used as totems for protecting recorded speech. Informally, 
\begin{enumerate}[topsep=-2.1pt, itemsep=-1pt, parsep=0.5ex]
    \item Consistency should be easy to verify and difficult to fake (described above);
    \item The totem-generated certificate $f_t(x)$ should be impacted by as many properties of the true signal $x$ as possible, so that a large portion of $x$ is protected.
\end{enumerate}

\myparagraph{Relation with cryptography. } The Totem framework has a cryptographical flavor, where the totem represents a function that is easy for the defender to work with (\eg, can invert), but hard to manipulate for the adversary. In fact, when each user is given a special totem, our framework is conceptually similar to cryptographical digital signature schemes, where a message is always sent together with a signature, used to verify message integrity and sender identity (\eg, the signature can be private-key encoding). However, unlike digital signatures, we need not design user-specific totems, and a totem holder can protect their signals even when a third party captures and transmits them.

\section{Method}
\label{sec:method}

For verification with the Totem framework, geometric consistency can be checked in various ways. Here we describe a specific procedure we use in this work, which verifies 3D consistency via scene reconstruction with neural radiance fields \cite{mildenhall2020nerf}. Specifically, the proposed method verifies the geometric consistency of a totem-protected image with the following 2 steps: (1) reconstructing the camera viewpoint from the provided totem views and (2) running a patch-wise comparison between the image and the reconstruction. 

Here we focus on spherical totems, which demonstrate the potential of this framework and avoid costly manufacturing of geometrically-complex totems. However, the method is not fundamentally limited in these aspects, as we discuss in details in  Section~\ref{sec:limitations}.

\subsection{Scene Reconstruction from Totem Views}\label{recon}
\textbf{Image formation process.} A totem-protected image is composed of image pixels $f(x)$ and distorted totem pixels $f_t(x)$. Image pixels capture the scene light rays directly passing through the camera optical system and display the scene as it appears to the naked eye. For totem pixels, the scene light rays first scatter through the refractive totems and then pass through the camera, implying that rays corresponding to two neighboring totem pixels may come from drastically different parts of the scene, depending on the complexity of the totem surface geometry. Regarding scene reconstruction, while traditional stereo methods suffice for simple totem views (\eg, radial distortion), they do not generalize to more distorted totem views. For this reason, we choose to model the scene as a neural radiance field \cite{mildenhall2020nerf} and reconstruct by rendering from the camera viewpoint.

\begin{figure*}[t]
\centering
\vspace{0.05in}
\begin{minipage}{.48\textwidth}
  \centering
  \vspace{0.15in} %
  \includegraphics[width=1.0\linewidth]{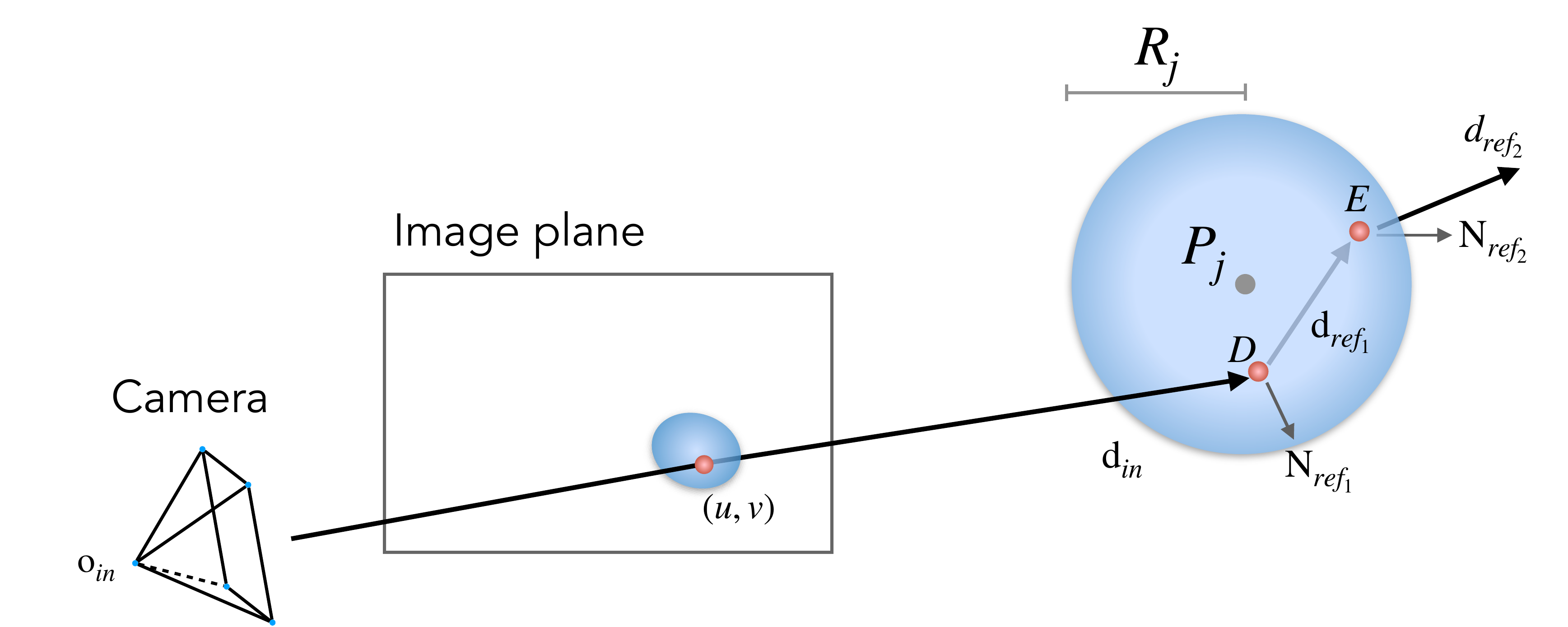}
  \vspace{-20pt}
  \vspace{0.2in} %
  \captionof{figure}{\small \textbf{Camera-ray refraction:} Using totem's geometrical properties, we compute the resultant ray direction of an image pixel that passes through the totem. For a spherical totem, this involves two refractions dependent on totem pose $P_j$.} 
  \label{fig:totem-refraction}
\end{minipage}
\hfill
\begin{minipage}{.48\textwidth}
  \centering
  \vspace{-0.05in}
  \includegraphics[width=1.0\linewidth]{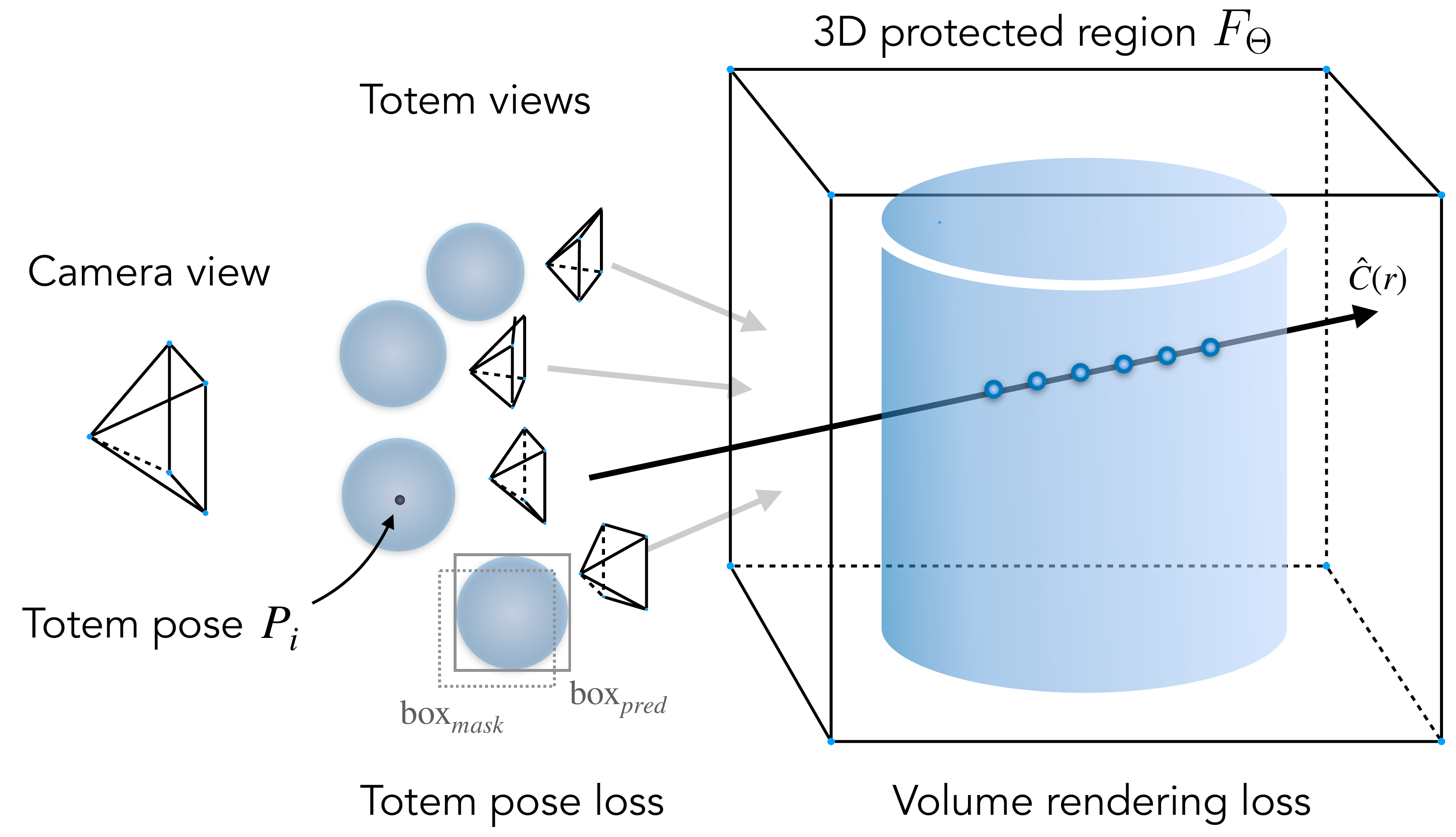}
  \vspace{-20pt}
  \captionof{figure}{\small \textbf{Reconstruction pipeline:} Using the refracted ray directions from the totems, we learn a radiance field $F_{\Theta}$ to reconstruct the 3D scene. We jointly optimize the totem positions $P_j$ with radiance field to improve the scene reconstruction.
  }
  \label{fig:totem-schematic}
\end{minipage}
\end{figure*}

A radiance field $F_{\Theta}$ represents a scene with a 5-dimensional plenoptic function that queries a spatial location and viewing direction and outputs its radiance and density. The color of an image pixel is rendered by querying 3D points along the corresponding scene light ray and computing the expected radiance given a distribution based on point density and occlusion. The mapping from image pixels to scene rays is simply the pinhole model, but for totem pixels, we need to compute the refracted ray which depends on totem-to-camera pose and totem properties such as 3D shape and index of refraction (IoR). With access to the totem 3D shape, we first register totem pose, obtain surface normals, and compute the mapping analytically. For simplicity, we ignore reflections on the totems and assume that the refraction process does not change light intensity.

\myparagraph{Pixel-to-ray mapping.} Given an image $\mathcal{I}$ and assuming a set of spherical totems $\mathcal{J}$ indexed $j=\{1 \ldots |\mathcal{J}|\}$, with center positions $P_j$ relative to the camera, radii $R_j$, and IoR $n_j$, and the index of refraction of air $n_{air}=1$, we first compute the  mapping from a totem pixel in the image $\mathcal{I}_{u, v}$ to the scene light ray corresponding to refraction through the totem $\mathbf{r}_{out} = \mathbf{o}_{out} + \mathbf{d}_{out} * t$, where $\mathbf{o}_{out}$ is the ray origin and $\mathbf{d}_{out}$ the ray direction. 

Following Fig.~\ref{fig:totem-refraction}, we begin with a ray $\mathbf{r}_{in} = \mathbf{o}_{in} + \mathbf{d}_{in} * t$ corresponding to pixel $(u, v)$ in the image, and compute point $D$ the first intersection of $\mathbf{r}_{in}$ with the totem, $\mathbf{N}_{ref_1}$ the surface normal at $D$, and $\mathbf{d}_{ref_1}$  the refracted direction:
\begin{align}
    D &= \mathsf{intersect}(P_j, R_j, \mathbf{d}_{in}, \mathbf{o}_{in}),\\
    \mathbf{N}_{ref_1} &= \overrightarrow{P_jD} / \|\overrightarrow{P_jD}\|_2 ,\\
    \mathbf{d}_{ref_1} &= \mathsf{refract}(n_j, n_{air}, \mathbf{N}_{ref_1}, \mathbf{d}_{in}).
\end{align}
\noindent Next, we compute the second intersection point $E$ where the ray exits the totem, corresponding surface normal $\mathbf{N}_{ref_2}$, and ray exit direction $\mathbf{d}_{ref_2}$:
\begin{align}
    E &= \mathsf{intersect}(P_j, R_j, \mathbf{d}_{ref_1}, D),\\
    \mathbf{N}_{ref_2} &= \overrightarrow{P_jE} / |\overrightarrow{P_jE}\|_2, \\
    \mathbf{d}_{ref_2} &= \mathsf{refract}(n_i, n_{air}, \mathbf{N}_{ref_2}, \mathbf{d}_{ref_1}).
\end{align}

\noindent We provide the formulas for $\mathsf{intersect}$ and $\mathsf{refract}$ in supplementary material. We obtain the resulting ray direction $\mathbf{r}_{out} = \mathbf{o}_{out} + \mathbf{d}_{out} * t$ with $\mathbf{o}_{out} = E$ and  $\mathbf{d}_{out} = \mathbf{d}_{ref_2}$. $\mathbf{r}_{out}$ is also refered to as $\mathbf{r}$ below.\\

\myparagraph{Joint optimization of radiance field and totem position.} A key part of the totem framework is determining the positions of the totem centers $P_j$ relative to the camera. While we know $P_j$ in simulator settings, it is necessary to estimate the totem positions in order to operate on real-world images. We assume that a binary mask $M_j$ for each totem is known, which could be annotated by the defender. We first initialize each $P_j$ by projecting the boundary pixels of the annotated binary mask into camera rays, forming a cone. Using the known totem radius $R_j$, we derive the totem's initial position by fitting a circle corresponding to the intersection of the cone and the spherical totem (details in supplementary).

We then jointly optimize the neural radiance function along with the totem position $P_j$. We use a photometric loss on $F_{\Theta}$ to reconstruct the color $\mathbf{C}(\mathbf{r})$ of totem pixels in the image corresponding to refracted totem rays $\mathbf{r}$ in batch $\mathcal{R}$ (see~\cite{mildenhall2020nerf} for construction of predicted color  $\hat{\mathbf{C}} (\mathbf{r})$): 
\begin{align}
    \mathcal{L}_{rec} = \sum_{\mathbf{r} \in \mathcal{R}} \left\Vert \hat{\mathbf{C}}  (\mathbf{r}) - \mathbf{C}  (\mathbf{r}) \right\Vert_2^2.
\end{align}
However, as minimizing per-pixel loss can cause $P_j$ to deviate far from the initial totem masks, we additionally regularize the optimized $P_j$ using an IoU loss. Given the current totem positions $P_j$ and radius $R_j$, we again compute the circle formed by the intersection of the totem with the cone of camera rays (see supplementary material for exact derivations). We sample a set of 3D points lying along this circle $\mathcal{X} = \{X_1 \ldots X_n\}$ and project them to 2D image coordinates using camera intrinsics $K$ and depth $d_i$: 
\begin{align}\label{eqn:perspective_projection}
    (u_i, v_i, 1) = \frac{KX_i}{d_i}.
\end{align}

We compute the bounding box of these projected 2D image coordinates: %
\begin{align}
    \mathrm{box}_{pred} = \left(\min_i(u_i),\; \max_i(u_i),\; \min_i(v_i),\; \max_i(v_i) \right).
\end{align}
We then apply the IoU loss~(Jaccard index~\cite{girshick2014rich}) over $\mathrm{box}_{pred}$ and $\mathrm{box}_{mask}$, the bounding box from the totem binary masks $M_j$, 
with the overall loss objective:
\begin{align}
    \mathcal{L} = \lambda  * \mathcal{L}_{rec}  + \mathcal{L}_{IoU},
\end{align}
\noindent where we use $\lambda = 10$. See additional training details in the supplementary.

\begin{figure*}[t]
    \centering
    \includegraphics[width=.95\textwidth]{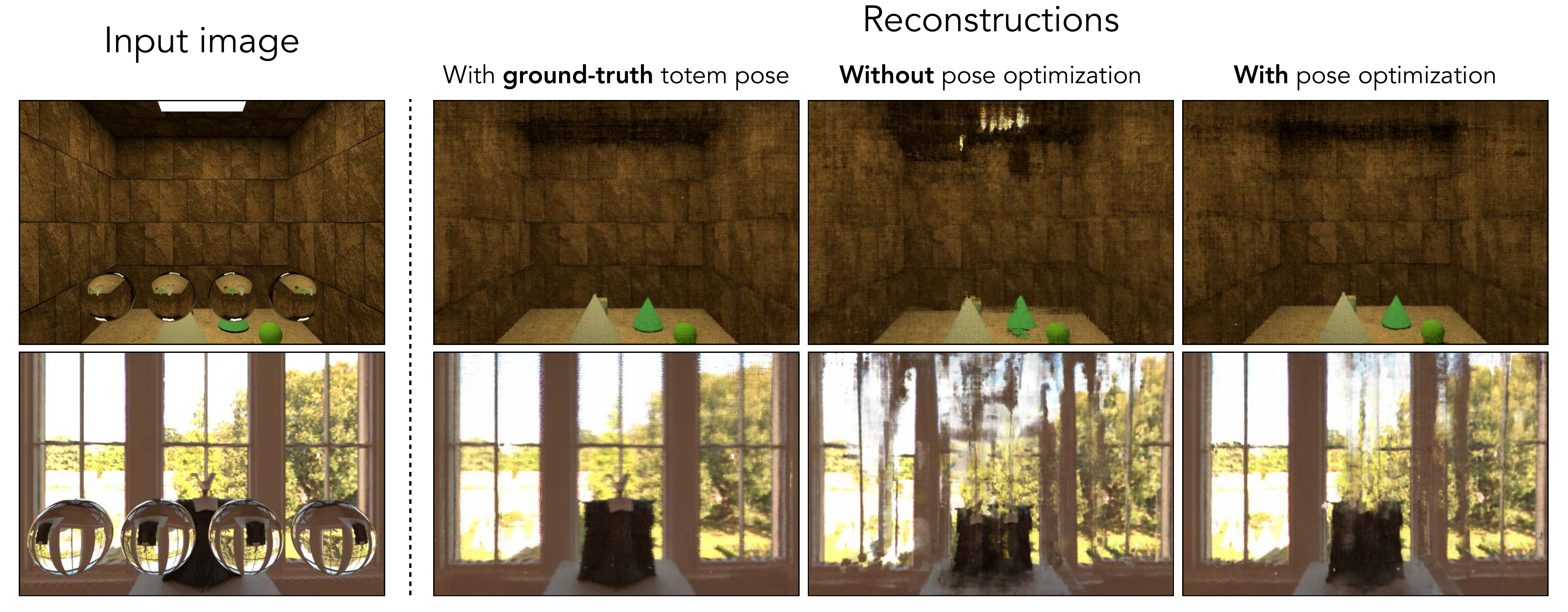}
    \caption[Pipeline]{\small \textbf{Pose optimization on synthetic data:} We start with a setup of spherical totems placed in a simulated environment. We find that four totems is sufficient to recover the scene when the totem pose is accurate; i.e., the ground-truth totem pose used to render the scene. Using the initial estimated totem position leads to artifacts in the reconstructed camera viewpoint, while allowing the totem poses to update while learning the scene representation improves the reconstruction.
    \label{fig:pose-results}}
\end{figure*}

\subsection{Manipulation Detection}\label{manip-detect}
As the totems do not uniformly cover the entire scene observed by the camera, we first construct a confidence map of the region intersected by multiple totems. For each totem ray $\mathbf{r}_{out}$, we sample points along the ray $\mathbf{r}_{out}(t) = \mathbf{o}_{out} + \mathbf{d}_{out} * t$ and query $F_{\Theta}$ to obtain their weight contribution $w(\mathbf{r}_{out}(t))$ to the resultant color. We then construct a 3D point cloud where each point $X_p$ is the point with the highest weight along a totem ray:
\begin{align}
    X_p = \mathbf{r}_{out}(\argmax_t w \left(\mathbf{r}_{out}(t) \right)).
\end{align}

We project these points to 2D using Eqn.~\ref{eqn:perspective_projection}. We accumulate the number of points from the point cloud that project to each pixel, apply a box filter of width 30 pixels, and threshold boxes with more than $10\%$ accumulated points. We take a convex hull around this thresholded region and call it the \textit{protected region}. Intuitively, this identifies the part of the scene that is adequately visible within the totems.

Within the protected region, we generate a heatmap for potential manipulations by comparing the scene reconstructed using the totems and $F_{\Theta}$ to the pixels visible in the image $\mathcal{I}$. We use a patch-wise L1 error metric:
{\small
\begin{align}\label{eqn:patchl1}
    \mathsf{L}_1(i,j) = \sum_{\substack{|k_i|<K\\|k_j| < K}} | \mathcal{I}(i+k_i, j+k_j) - \hat{\mathbf{C}}(i+k_i, j+k_j) |,
\end{align}%
}
with patch size $K = 64$, where $\hat{\mathbf{C}}(i, j)$ refers to the color along the camera ray corresponding to pixel $(i, j)$. In addition to patch-wise L1, we also use LPIPS, a learned perceptual patch similarity metric~\cite{zhang2018unreasonable}, on the same patch size.

\begin{figure*}[t]
    \centering
    \includegraphics[width=1\textwidth]{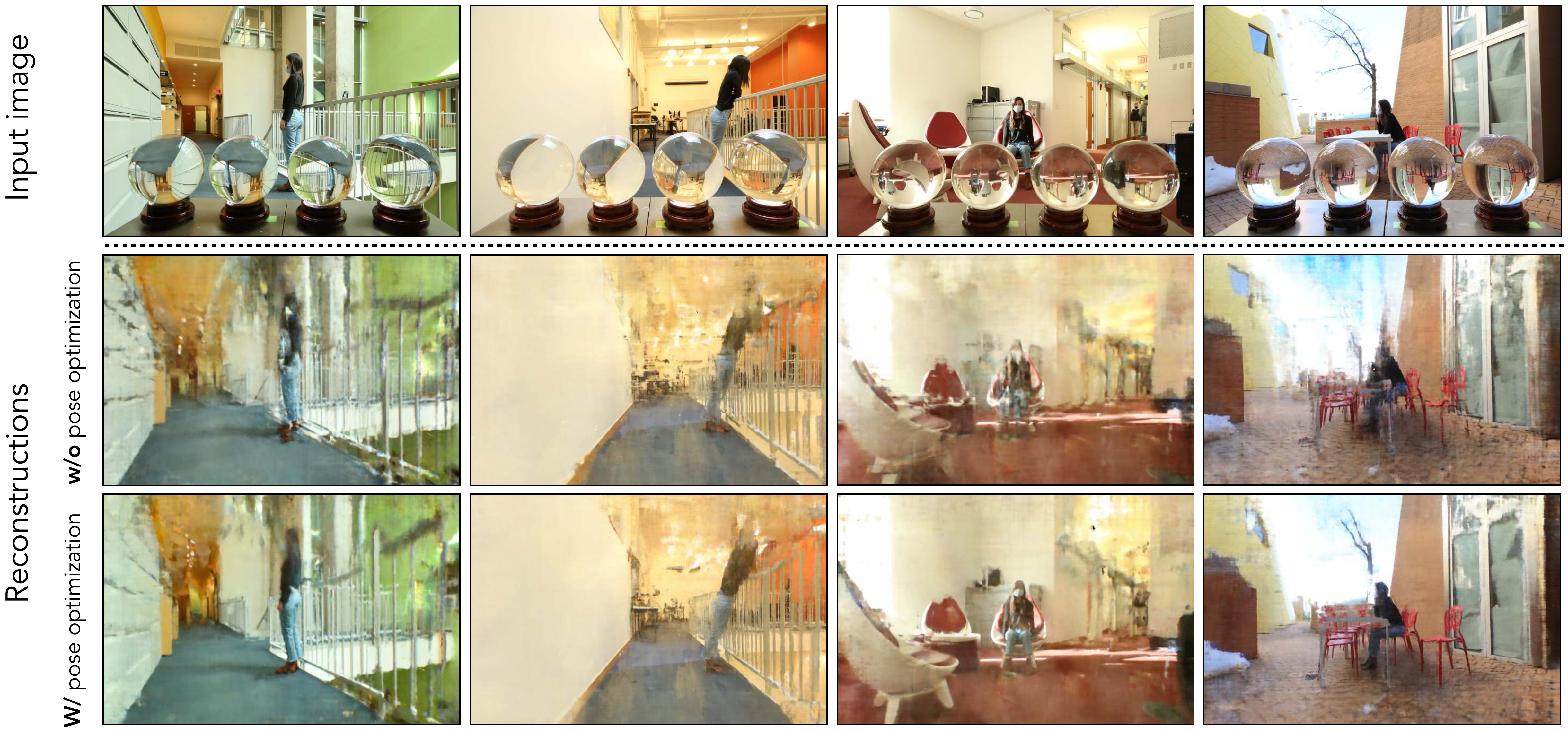}\vspace{-7pt}
    \caption[Pipeline]{\small \textbf{Reconstruction results on real images:} For \textit{real images}, we do not know the ground-truth totem positions and must rely on our position estimates. We find that jointly optimizing the pose of the totems together with the scene reconstruction better recovers the scene geometry and obtains a closer match to the input image than using the initial totem pose estimates. We conduct reconstruction on indoor and outdoor scenes under a range of lighting conditions.
    \label{fig:reconstruction-results}}
\end{figure*}

\begin{figure*}[h]
    \centering
    \includegraphics[width=1.0\textwidth]{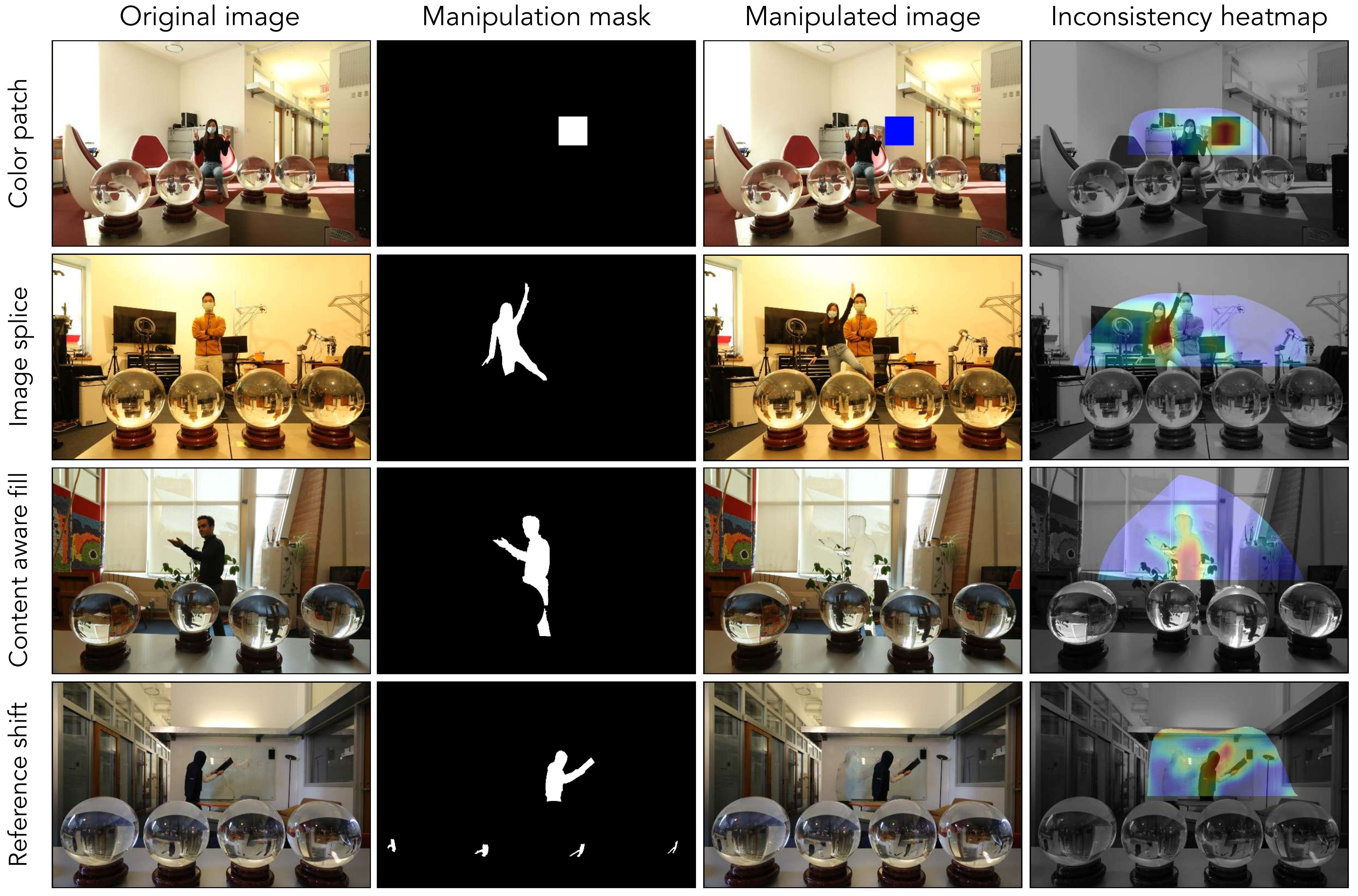}\vspace{-7pt}
    \caption[Pipeline]{\small \textbf{Detection results:} We manipulate scenes by adding random color patches, inserting people with image splicing, removing people with Photoshop (CAF), and shifting people in both camera and totem views to the same reference point (\eg right edge of the wooden table). By comparing the manipulated image and the scene reconstruction, we obtain an inconsistency heatmap over regions of possible manipulation.
    \label{fig:detection}}
\end{figure*}

\section{Results}\label{sec:results}

\subsection{Data Collection}

\myparagraph{Synthetic images.} We first demonstrate our method in a simulated setting, where we know all ground-truth information about the totems and camera. We generate the data with Mitsuba2~\cite{nimier2019mitsuba}, a differentiable rendering system. We try two settings: (1) we set up a room similar to a Cornell box with random wallpapers and random geometric objects on the floor (2) we generate the room using an environment map (Fig.~\ref{fig:pose-results}-left). We then place refractive spheres in between the camera and the scene to form the totem views.

\myparagraph{Real images.} To demonstrate our framework in more realistic settings, we take pictures using a Canon EOS 5D Mark III camera and place four physical totems at arbitrary positions in front of the scene (Fig.~\ref{fig:detection}-left). We obtain the totem size and IoR from manufacturer specifications and manually annotate the totem masks in the image. Camera intrinsic parameters are obtained via a calibration sequence. We correct radial distortion in the collected images to better approximate a pinhole camera model when computing refracted ray directions.

\myparagraph{Image manipulations.} We conduct manipulations to create inconsistencies between the observed scene and the totem views. We locally modify the image by inserting randomly colored patches,  adding people by image splicing, removing people with Photoshop Content Aware Fill (CAF), or shifting people in both image and totems to the same reference position. Note that we do not consider manipulations where totems are entirely removed; in that case we consider the image no longer verifiable.

\subsection{Decoding the Scene from Totem Views}

\myparagraph{Reconstructing a simulated scene.} We conduct initial experiments in simulated scenes to validate components of our learning framework when ground-truth totem parameters are known. Fig.~\ref{fig:pose-results} shows the reconstruction of the scene using (1)  the known totem positions for reconstruction as the oracle (2) only the initial estimate of totem positions derived from annotated totem masks, and (3) jointly estimating the totem position and scene radiance field. We find that small changes in the totem position greatly impact the reconstruction quality; therefore relying on the initial totem position estimate alone leads to sub-optimal reconstruction of the camera viewpoint. The reconstruction improves when allowing the totem positions to update during learning, while using the oracle ground truth totem position obtains the best reconstruction (Tab.~\ref{tab:recon-table}). However, ground truth totem positions are only available in simulators, so we must estimate these positions when using real images. In supplementary material, we conduct additional experiments on the number of totems required to reconstruct the scene and find empirically that four totems leads to a reasonable balance between reconstruction quality and visibility of the scene.  %

\myparagraph{Reconstructing scenes from real images.} 
Similar to the simulated environment, we set up four totems in a room in front of the subject, and jointly optimize
for reconstruction and totem position. We find that the joint optimization procedure yields a better scene reconstruction when viewed from the camera (Fig.~\ref{fig:reconstruction-results}). On un-manipulated images, joint optimization decreases the L1 error of the reconstructed scene to the ground truth from 0.15 to 0.11 (Tab.~\ref{tab:recon-table}). %

\myparagraph{Detecting image manipulations.} We next investigate the ability to detect manipulations after reconstructing the scene from only the totem views. We experiment with a patch-wise L1 distance metric (Eqn~\ref{eqn:patchl1}) and a perceptual distance metric~\cite{zhang2018unreasonable} to measure the difference between the camera viewpoint and the scene reconstruction, yielding a heatmap over the potentially manipulated area. Qualitative examples are shown in Fig.~\ref{fig:detection} and we quantify the detection performance in Tab.~\ref{tab:detection-table} by normalizing the heatmap and computing average precision over these patches. 
While our method relies on 3D geometric consistency obtained from a single image, we compare to an image splice detection method~\cite{huh2018fighting}, but we note that such learning-based methods tend to fail on setups outside of the training distribution and where manipulations involve parts of two images with the same camera metadata. We also compare to Wu et al. ~\cite{Wu2019ManTraNet} with downsampled images due to GPU memory explosion. The low final mAP is partly due to the method's sensitivity to the exact compression artifacts, which can be affected by any processing (e.g., resizing).

\begin{table}[t]
  \centering
  \begin{minipage}{.48\textwidth}
  \scalebox{0.8}{
  \begin{tabular}{lcccc}
    \toprule
    \multirow{2}{*}{Dataset} & \multirow{2}{*}{\shortstack[c]{Totem \\ Optimization}}  & \multicolumn{2}{c}{Reconstruction} & Pose \\
    \cmidrule(lr){3-4} \cmidrule(lr){5-5}
     &  & L1 & LPIPS & L1 \\
    \midrule
    Box & \xmark &  0.057 & 0.658 & \textbf{0.008} \\
    Box & \checkmark &  0.054 & 0.645 & 0.108 \\
    Box & Oracle &  \textbf{0.047} & \textbf{0.625} & - \\
    \midrule
    Env map & \xmark &  0.173 & 0.617 & 0.060 \\
    Env map & \checkmark &  0.103 & 0.520 & \textbf{0.027} \\
    Env map & Oracle &  \textbf{0.040} & \textbf{0.476} & - \\
    \midrule
    Real & \xmark &  0.149 & 0.644 & - \\
    Real & \checkmark &  \textbf{0.109} &\textbf{ 0.586} & - \\
    \bottomrule
  \end{tabular}
  }%
  \vspace{0.2in}
  \caption{\small \textbf{Camera view reconstruction comparisons:} We measure L1 and LPIPS~\cite{zhang2018unreasonable} distance of the camera view reconstruction using the learned scene representation. 
  \label{tab:recon-table}}
  \end{minipage}
\hfill
  \begin{minipage}{.48\textwidth}
      \scalebox{0.8}{
  \begin{tabular}{llccc}
    \toprule
    \multicolumn{2}{c}{Method} & CAF & Splice & Color \\
    \midrule
    \multicolumn{2}{c}{Self-consistency~\cite{huh2018fighting}} & 0.037 & 0.037 &  0.801\\[0.6ex]
    \multicolumn{2}{c}{ManTra-Net~\cite{Wu2019ManTraNet}} & 0.181 & 0.151 &  0.295\\[0.6ex]
    \multirow{2}{*}{\shortstack[c]{Ours w/o totem opt.}} & +L1 & 0.485 & 0.401 &  0.944 \\
    & +LPIPS & 0.489 & 0.449 &  0.954\\[0.6ex]
    \multirow{2}{*}{\shortstack[c]{Ours with totem opt.}} & +L1 & 0.554 & 0.638 &  \textbf{0.961}\\
    & +LPIPS  & \textbf{0.666} & \textbf{0.739} &  0.946\\
    \bottomrule
  \end{tabular}
  }%
  \vspace{0.2in}
  \caption{\small \textbf{Detection comparisons:} Patch-wise mAP on various image manipulations. 
Compared to  \cite{huh2018fighting} and \cite{Wu2019ManTraNet}, our method is based on geometric reconstructions and is therefore robust to different manipulation types and image processing.
  \label{tab:detection-table}}

  \end{minipage}

\end{table}

\subsection{Potential Avenues for a More General Method}

\label{sec:limitations}

In our current method as described in Section~\ref{sec:method}, we make several simplifying assumptions about the totems and scene. These assumptions are not fundamental limitations of the framework, as they can potentially be addressed by, for example, high-precision totem manufacture/measurement, advanced reconstruction method not requiring known camera intrinsics and/or robust to more diverse totem placements, etc.  We briefly discuss these limitations below. Please refer to the FAQ in the supplementary materials for more discussions.

\myparagraph{Reconstruction.} For high reconstruction quality, the current method is best suited when scene is clearly visible in multiple totem views, so that the neural radiance field fitting has more training samples (pixels) and is more stable. Therefore, detection is more difficult when totem placements only show the manipulated parts in a rather small view  (e.g., totems far away from camera) or a highly distorted view (e.g., totems near the sides of the image). See the supplementary material for examples and analysis on totem placement (including number of totems and totem positions). Based on neural radiance fields, the reconstruction is also not fast enough for real-time detection and assumes known camera intrinsics.
As scene reconstruction research continues to develop, we believe these limitations will be become much less relevant.

\myparagraph{Totem design.} %
Our experiments use spherical totems. They can be readily purchased online and have known geometry, which allow us to much more easily experiment and analyze this new verification strategy.
However, our verification framework, specifically the reconstruction component, is not limited to this one geometry. It directly generalizes to various totem designs as long as the totem geometry and physical properties can be computed/measured. For example, Zhang et al.~\cite{zhang2015sparklevision} demonstrates reconstructions under a complex scrambling of light rays once the input-output ray mapping is known.

Much more can be explored in designing totems for ease to use and effectiveness towards forensics tasks.
For example, totems can be made more compact, and therefore more portable and less visible to the adversary (totem identities are unknown to the adversary a priori). Totems with complex geometric design exhibit less interpretable distortion patterns and can contain multiple distorted views of the scene simultaneously, which makes it more difficult to achieve geometrically-consistent manipulation and reduces the number of totems required during scene setup.

\section{Conclusion}

We design a framework for verifying image integrity by placing physical \textit{totems} into the scene, thus encrypting the scene content as a function of the totem geometry and material. By comparing the scene viewed from the camera to the distorted versions of the scene visible from the totems, we can identify the presence of image manipulations from a single reference image. Our approach decodes the distorted totem views by first estimating the totem positions, computing the refracted ray directions, and using the resultant rays to fit a scene radiance field. Furthermore, we show that it is possible to fit this 3D scene representation using sparse totem views, and that jointly optimizing the totem positions and the scene representation improves the reconstruction result. While we assume spherical totems in this work, an avenue for future exploration would be to extend the approach to more complex totems such as those with more complex shapes or randomly oriented microfacets, thus creating a stronger encryption function. \\

\subsubsection{Acknowledgements.} This work was supported by Meta AI. We thank Wei-Chiu Ma, Gabe Margolis, Yen-Chen Lin, Xavier Puig, Ching-Yao Chuang, Tao Chen, and Hyojin Bahng for helping us with data collection. LC is supported by the National Science Foundation Graduate Research Fellowship under Grant No. 1745302 and Adobe Research Fellowship.

\newpage
\bibliographystyle{splncs04}
\bibliography{egbib}

\newpage
\begin{center}
    \Large
    \textbf{Supplementary Material}
    \vspace{0.4cm}
\end{center}
\renewcommand\thesection{\Alph{section}}
\setcounter{section}{0}

In this document, we first discuss additional implementation details regarding ray refraction operations and training of the radiance field (Section~\ref{sec:sm_method}). We conduct additional experiments investigating the number and configuration of totems, and show results on more scenes in Section~\ref{sec:sm_experiments}. We address different modes of image manipulation and provide a brief FAQ in Section~\ref{sec:sm_scope}.

\section{Additional method details}\label{sec:sm_method}

\subsection{Pixel-to-ray mapping} 
\textbf{Overview.} Recall that the first step of our method is to infer the underlying 3D scene using the refracted rays corresponding to the distorted totem pixels. For a given image $\mathcal{I}$ and a set of spherical totems $\mathcal{J}$ indexed $j=\{1 \ldots |\mathcal{J}|\}$, with center positions $P_j$ relative to the camera, radii $R_j$, and IoR $n_j$, we compute the mapping from a totem pixel in the image $\mathcal{I}_{u, v}$ to the scene light ray corresponding to refraction through the totem $\mathbf{r}_{out} = \mathbf{o}_{out} + \mathbf{d}_{out} * t$. We decompose this mapping procedure into two steps: 
\begin{enumerate}
  \item Begin with a ray $\mathbf{r}_{in} = \mathbf{o}_{in} + \mathbf{d}_{in} * t$ corresponding to pixel $\mathcal{I}_{u, v}$. Compute the first intersection $D$ with totem $j$ and the direction $\mathbf{d}_{ref_1}$ of the refracted ray entering the totem.
  \item Take the intermediate ray $\mathbf{r}_{mid} = \mathbf{o}_{mid} + \mathbf{d}_{mid} * t$, where $\mathbf{o}_{mid} = D$ and $\mathbf{d}_{mid} = \mathbf{d}_{ref_1}$. Compute the second intersection $E$ with totem $j$ and the  direction $\mathbf{d}_{ref_2}$ of the refracted ray exiting the totem.
\end{enumerate}

In Sec.~\ref{recon} in the main text, we use a general $\mathsf{intersect}$ function to compute the ray-totem intersections $D$ and $E$ and a general $\mathsf{refract}$ function to compute the refracted ray directions $\mathbf{d}_{ref_1}$ and $\mathbf{d}_{ref_2}$. Below we provide the formula and implementation details for these two functions.\\

\myparagraph{Define} $\mathsf{intersect}$. Given a ray $\mathbf{r} = \mathbf{o} + \mathbf{d}  * t$ and a sphere with radius $R$ and center position $P$, the $\mathsf{intersect}$ function first confirms the validity of the ray-sphere intersection, then computes the two intersection positions and returns the one closest to the ray origin $\mathbf{o}$ and in the ray direction $\mathbf{d}$. Since an intersection $X$ must satisfy both the sphere equation $\lVert X - P \rVert_2^2 = R^2$ and the ray equation $X = \mathbf{o} + \mathbf{d}  * t$, we formulate the $\mathsf{intersect}$ function as the optimization below:
\begin{align}
\mathsf{intersect}(P, R, \mathbf{d}, \mathbf{o}):= \mathbf{o} + \mathbf{d}  * \left( \argmin_t \left | R^2 -  \lVert \mathbf{o} + \mathbf{d}  * t - P \rVert_2^2  \right | \right).
\end{align}

To solve for $t$, we set the inner optimization term to $0$ and use a quadratic solver  $\mathsf{quad}\left(a, b, c\right)$ with input arguments $a = \|\mathbf{d}\|_2^2$, $b = 2\langle \mathbf{o}-P, \mathbf{d}\rangle$, $c = \|\mathbf{o}-P\|_2^2$. Before this step, we confirm that the input ray has valid intersections with the sphere by checking if the discriminant term in the quadratic solution is positive.\\

\myparagraph{Define} $\mathsf{refract}$. The function $\mathsf{refract}$ computes the refracted direction $\mathbf{d}_{ref}$ of an incident ray $\mathbf{d}_{in}$. Given the unit surface normal $\mathbf{N}$ at the ray-medium intersection, IoR of the incident medium $n_1$ and the refractive medium $n_2$, we derive an analytical solution using the Snell's law:
\begin{align}
    \mathsf{refract}(n_1, n_2, \mathbf{N}, \mathbf{d}_{in}):= 
    \frac{n_1}{n_2} \left(\mathbf{N} \times (-\mathbf{N} \times \mathbf{d}_{in})\right) - \mathbf{N} \sqrt{1-\frac{{n_1}^2}{{n_2}^2} \lVert \mathbf{N} \times  \mathbf{d}_{in} \rVert_2^2 }.
\end{align}

\subsection{Totem pose} \label{totem-pose}

\begin{figure*}[h]
\centering
\includegraphics[width=\linewidth]{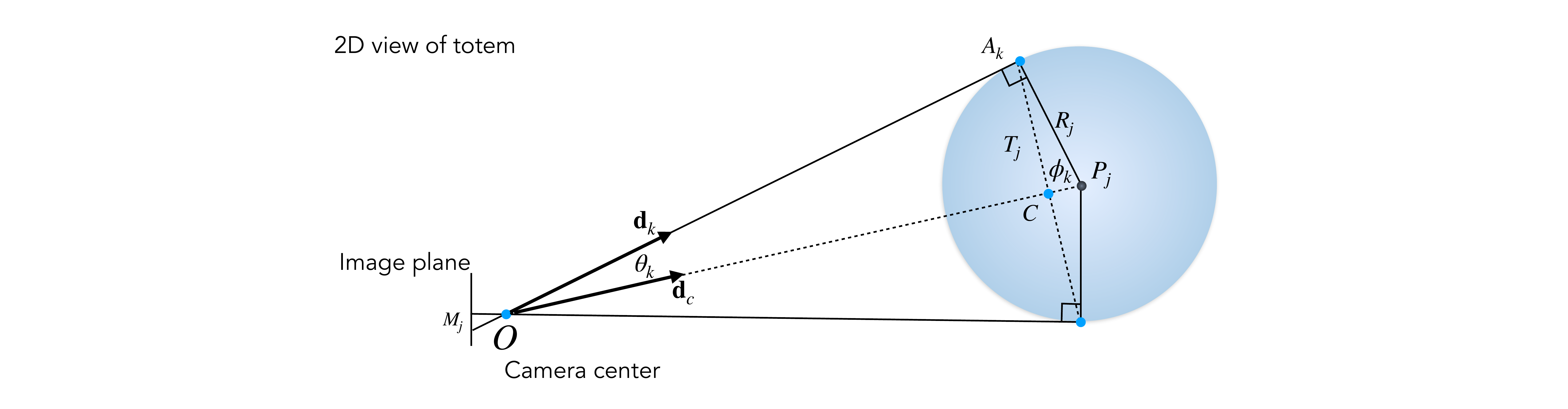}\vspace{-2.5ex}
\caption{2D visualization of the intersection between the spherical totem $j$ and tangent rays $\mathbf{r}_k = O + \mathbf{d}_k  * t$ corresponding to the boundary pixels of totem mask $M_j$. This intersection forms a circle in 3D space with radius $T_j$ and center $C$. We use this circle to obtain an initial estimate of the totem center $P_j$ and also use its 2D projection on the image plane to regularize the totem position during joint optimization by enforcing consistency with the provided totem mask $M_j$. \label{fig:supp-totem-cone}}
\end{figure*}

\noindent \textbf{Totem pose initialization.}
To obtain an initial estimate of the totem positions, we use an optimization procedure to fit the boundary of the mask pixels $M_j$ to a circle that corresponds to the intersection between the spherical totem $j$ and the tangent cone formed by the boundary rays (\Cref{fig:supp-totem-cone}). 

First, assuming the camera center $O$ is at the origin, we express the rays corresponding to the boundary pixels of a totem mask $M_j$ as $\mathbf{r}_k = \mathbf{d}_k  * t$ and normalize $\mathbf{d}_k$ to have unit length. When projected into 3D space, these $K$ rays form a tangent cone with the totem, and we estimate the cone axis by averaging the unit-length boundary vectors:
\begin{equation}
\mathbf{d}_c = \frac{1}{K} \sum_k{\mathbf{d}_k}.
\end{equation}
We then normalize $\mathbf{d}_c$ to unit length and solve for the angle between $\mathbf{d}_k$ and $\mathbf{d}_c$:
\begin{equation}
\theta_k = \arccos (\mathbf{d}_k \cdot \mathbf{d}_c ).
\end{equation}

With known totem radius $R_j$ and given that the tangent ray $\mathbf{r}_k$ is perpendicular to $\overrightarrow{P_j A_k}$, where $A_k$ is the point of tangency, we solve for the complementary angle $\phi_k = \frac{\pi}{2} - \theta_k$ and estimate the radius of the circular intersection $T_j = \frac{1}{K} \sum_k R_j \sin (\phi_k)$. Next, we solve for the slant height $t_{est}$ of the tangent cone by minimizing the objective function:
\begin{equation}
    t_{est} = \argmin_t \bigg| \frac{1}{K} \sum_k{||\mathbf{d}_k  * t - C||}  - T_j \bigg|,
\end{equation}
where $C$ is the cone base center expressed as $C = \frac{1}{K} \sum_k{\mathbf{d}_k  * t}$. Intuitively, the boundary rays of a totem mask $M_j$ defines a cone with a fixed opening angle and we optimize the slant height $t_{est}$ for this cone such that the cone radius matches the previously solved radius $T_j$.
Using the estimated $t_{est}$, we then compute the estimated totem center $P_j$ step by step:
\begin{align}
    C &= \frac{1}{K} \sum_k{\mathbf{d}_k  * t_{est}}\\
    |\overrightarrow{OC}| &= ||C||\\
    |\overrightarrow{P_j C}| &= \frac{T_j^2}{|\overrightarrow{OC}|} \quad \text{(similar triangles)}\\
    |\overrightarrow{P_j O}| &= |\overrightarrow{P_j C}| + |\overrightarrow{OC}|\\
    P_j &= \mathbf{d}_c  * |\overrightarrow{P_jO}|.
\end{align}

Due to inaccuracies in the totem masks, particularly for real images in which the totems are manually segmented, we note that the above procedures involve a number of approximations. Thus, we find that using this as an initialization and further refining the totem positions yields better reconstruction results. 

~\\ \noindent \textbf{Totem IoU Loss.} To regularize the totem positions during optimization, we use an IoU loss between the bounding box extracted from the totem mask and the bounding box estimated from the totem position during training. To obtain the latter bounding box, we reverse some of the calculations above and start with the current totem position $P_j$ and camera origin $O$ to obtain segment $|\overrightarrow{P_jO}|$; together with known radius $R_j$, we solve for the circle radius $T_j$ and cone center $C$ using similar triangles. 
We obtain the normal vector of the circular intersection as:
\begin{equation}
    \mathbf{n} = \frac{P_j - C}{||P_j - C||}.
\end{equation}

With the normal vector $\mathbf{n}$, center $C$, and radius $T_j$, we have a defined 3D circle. Next, we evenly sample $N=1000$ points along the circle and project them onto the image plane using perspective projection. Taking the minimum and maximum x and y coordinates of these projected points yields the bounding box used for the IoU loss.

\subsection{Radiance field training}
\textbf{Pre-processing.} We describe two pre-processing steps for improving reconstruction quality. First, for each scene light ray $\mathbf{r}_{out} = \mathbf{o}_{out} + \mathbf{d}_{out}  * t$ computed from a totem pixel $\mathcal{I}_{u, v}$ (Sec.~\ref{recon} main text), we shift $\mathbf{o}_{out}$ to the ray's intersection with the plane $z=0$ and scale $\mathbf{d}_{out}$ to have unit length in the $z$ direction. Next, we map the normalized rays from camera space to a cube space $[-1, 1]^3$ and filter out rays if the mapped ray origins fall outside of the cube space by a threshold. This automatically removes rays with large refraction angles. These rays can make training unstable, as small updates to the totem positions result in large changes in refracted ray directions.

\myparagraph{Training details.} We first train the neural radiance model alone for 100 epochs and then jointly optimize with the totem positions for another 49.9k epochs. Training takes approximately 5 hours on one NVIDIA GeForce RTX 2080 Ti. For the neural radiance model, we follow the same training and rendering procedures in Mildenhall \textit{et al.} ~\cite{mildenhall2020nerf}. During joint optimization, instead of estimating the absolute totem positions, we learn the relative translation from initial totem positions obtained in Sec.~\ref{totem-pose}. For training the totem parameters, we use the Adam optimizer with a learning rate of $0.00001$ and scales the learning rate with $\gamma = 0.99$ every 100 epochs.

\section{Additional experiments}\label{sec:sm_experiments}

\subsection{Number of totems}
\begin{figure}[h]
     \centering
     \includegraphics[width=\textwidth]{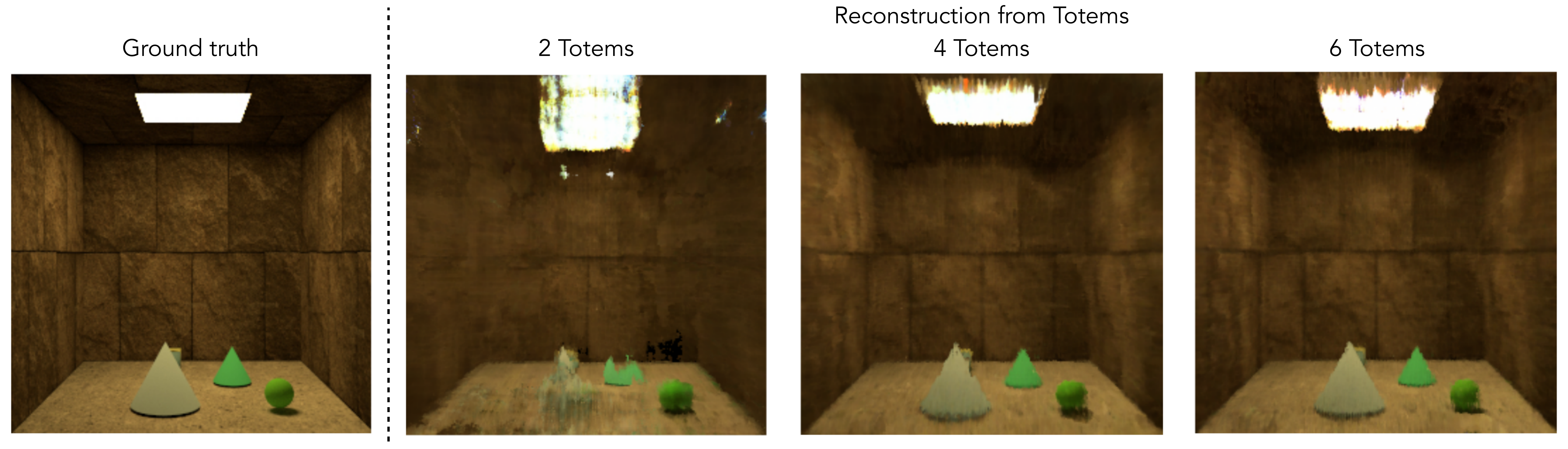}
    \caption{\textbf{Number of totems.} Reconstructions obtained by varying the number of totems in the scene. We find that four totems is sufficient to obtain a reasonable reconstruction, while also balancing the visibility of the background scene. 
    \label{fig:howmany}}
\end{figure}

We experiment with placing different numbers of totems in a simulated scene in Fig.~\ref{fig:howmany}. While using two totem views results in a poor reconstruction of the scene, the result improves when using four or six totems. Quantitatively, we find that using two totems yields the worst reconstruction error (0.16 L1 error), and using four and six totems attains better reconstruction error with four totems being slightly better (0.08 and 0.09 respectively). %
We note that while the reconstructions from four and six totems are also qualitatively similar, placing more totems results in more occlusion of the actual scene. Therefore, we chose to proceed with a four-totem scene setup in further experiments.

\subsection{Patch-level detection}

In addition to detecting image-level manipulation, here we detect whether each patch of an image is manipulated and provide new quantitative measures and visualizations to demonstrate the performance of our detection method.

\begin{figure}[h]
\centering
\includegraphics[width=\linewidth]{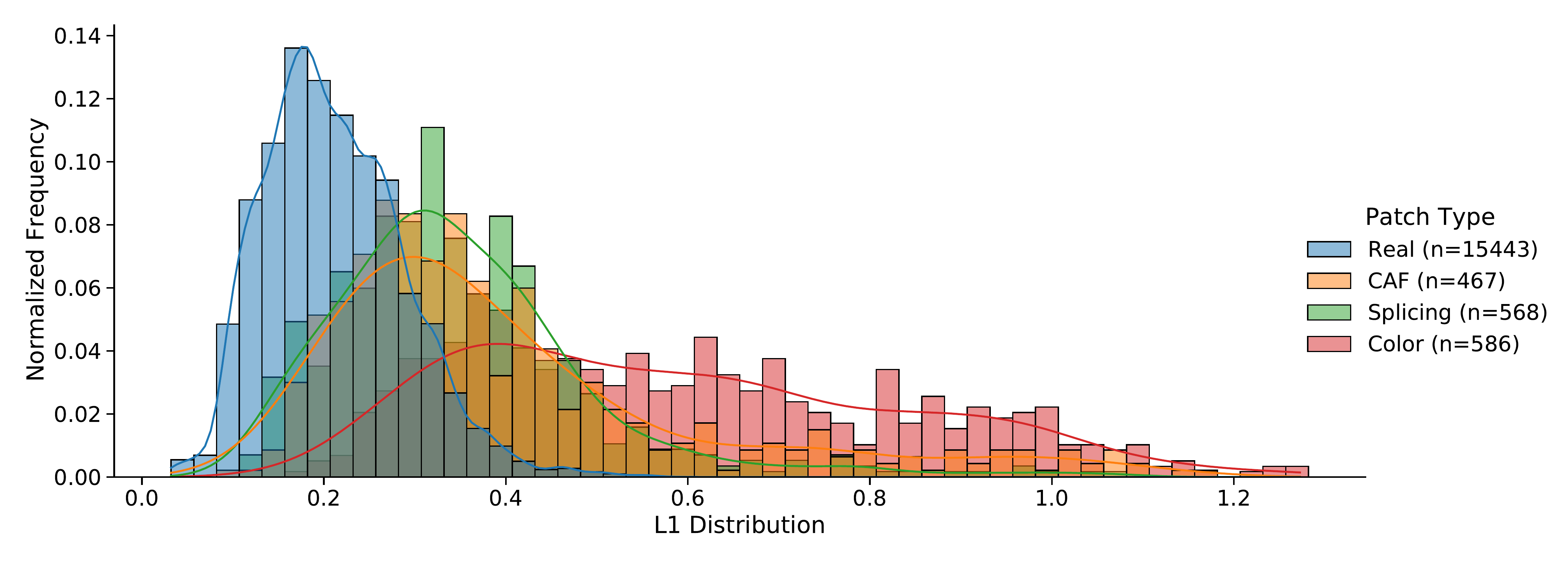}\\[-2.5ex]
\includegraphics[width=\linewidth]{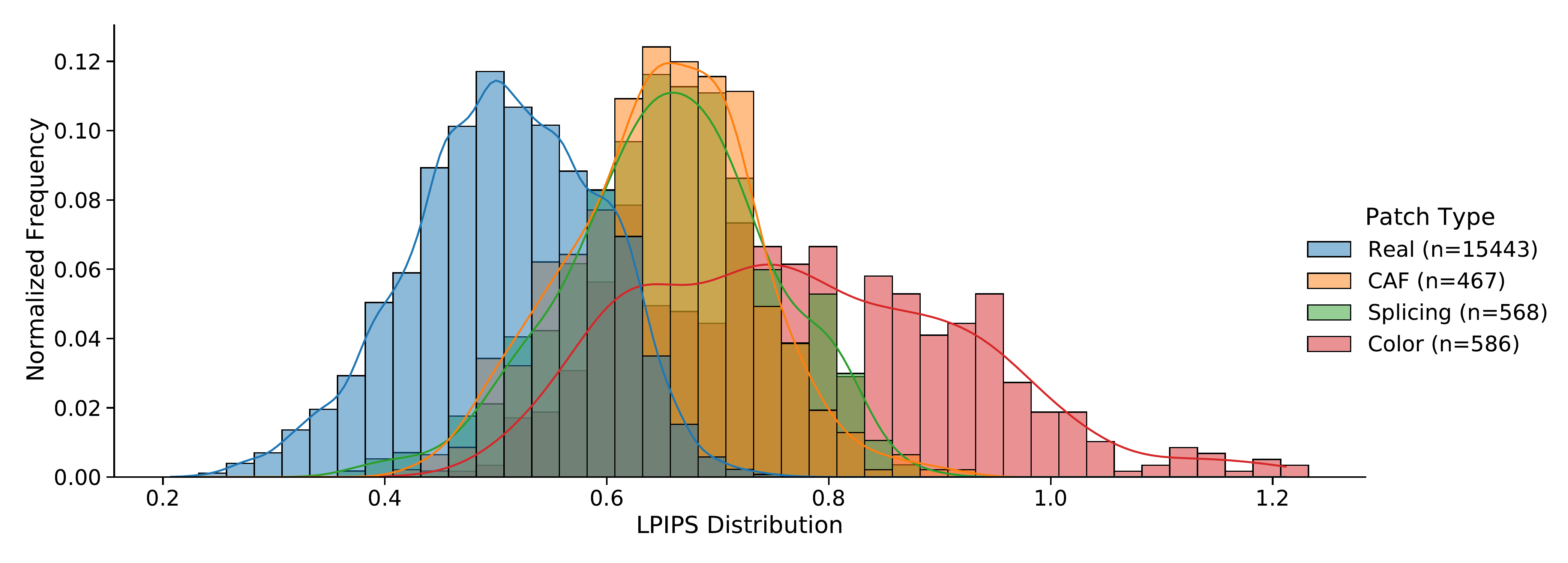}\vspace{-2.5ex}
\caption{Distributions of L1 and LPIPS metrics on \emph{patches} (unmanipulated \emph{real} patches and patches manipulated in different ways). For both metrics, our method  overall gives real patches lower scores than manipulated ones.} \label{fig:supp-patch-score-distn}
\end{figure}

To remain consistent with Tab.~\ref{tab:detection-table} in the main paper, we use the same $37$ images, including $7$ unmaniplated images, $7$ CAF images, $8$ spliced images, and $15$ images with added color patches. For each image, we extract $900$ patches of $64\times64$ resolution, over a $30\times30$ grid evenly spaced horizontally and vertically above the totem area. After extraction, only patches that overlap with corresponding protect regions are kept. This results in a total of $17064$ patches, of which $1621$ are manipulated (that is, having $> 10\%$ manipulated pixels). The exact number of patches for each manipulation can be found in \Cref{fig:supp-patch-score-distn}. For each patch, we compute L1 or LPIPS against the corresponding patch from the reconstructed view.

In \Cref{fig:supp-patch-score-distn}, we visualize the distribution of our two metrics (L1 and LPIPS) for each type of patches. For both metrics, the distributions exhibits a qualitative difference between real patches and manipulated ones, giving an overall lower score to real patches. Indeed, our metrics help detecting manipulations. In \Cref{tab:supp_patch-score-ap}, they lead to nontrivial gains in terms of average precision. Note the imbalance between the numbers of real and manipulated patches.

\begin{table}
  \centering
  \scalebox{0.85}{
  \begin{tabular}{llcccc}
    \toprule
    & %
    & \multirow{2}{*}{\shortstack[c]{CAF+Real\\($7.26\%$ manip.)}} 
    & \multirow{2}{*}{\shortstack[c]{Splice+Real\\[-0.4ex]($8.20\%$ manip.)}} 
    & \multirow{2}{*}{\shortstack[c]{Color+Real\\($5.78\%$ manip.)}} 
    & \multirow{2}{*}{\shortstack[c]{All Patches\\($9.50\%$ manip.)}} \\
    & & & & & \\
    \midrule
    \multirow{2}{*}{\shortstack[c]{Ours with totem opt.}} %
    & +L1     
    & 0.4412 & 0.5026 & 0.8554 & 0.6455 \\
    & +LPIPS  
    & 0.4954 & 0.6315 & 0.8169 & 0.7086 \\
    \midrule 
    \multicolumn{2}{c}{Na\"{i}ve Detector (Random Decision)} %
    & 0.0726 & 0.0820 & 0.0578 & 0.0950 \\
    \bottomrule
  \end{tabular}
  } %
  \vspace{0.2in}
  \caption{\textbf{Patch-level detection comparisons:} Average precision of our method on patches created with different types of images. For example, CAF+Real contains all patches from CAF images and unmanipulated images, while the last column (All Patches) shows results on all patches from all images. To show class imbalance, we also report the precision of a na\"{i}ve detector that randomly detects its output, which equals the ratio of manipulated patches.
  \label{tab:supp_patch-score-ap}}
\end{table}

\begin{figure}[t]
    \centering
    \includegraphics[width=1.0\linewidth]{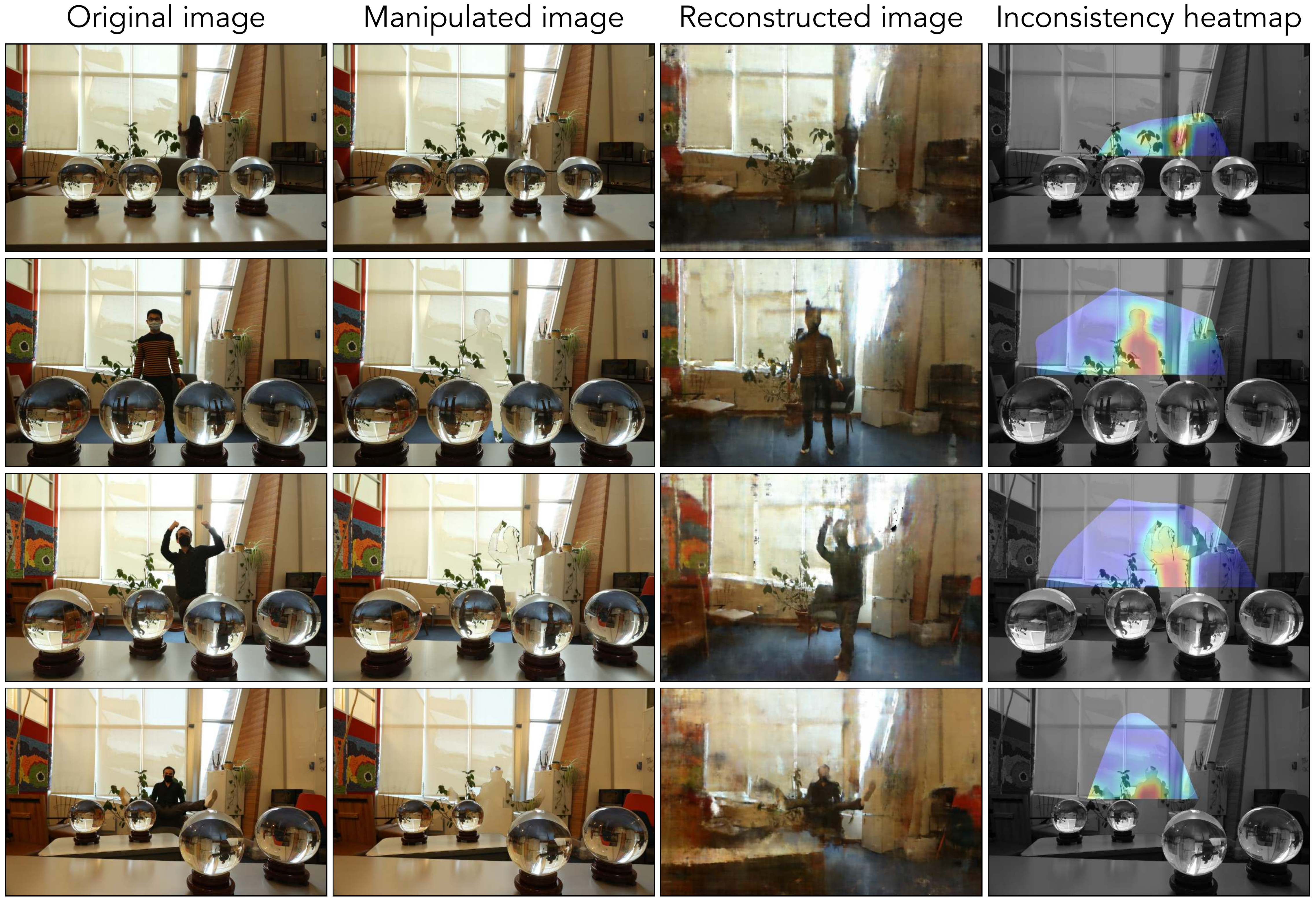}
    \vspace{-7pt}\caption[]{\textbf{Comparison of different totem configurations.} Additional reconstruction and detection results for the same scene while varying the subject and totem configurations. We manipulate all above scenes by removing people with Photoshop Content Aware Fill. Our method has consistent reconstruction quality and detection results under various totem configurations.} 
    \label{fig:supp-results}
\end{figure}

\begin{figure}
\centering
\vspace{-7pt}
\includegraphics[width=1.0\linewidth]{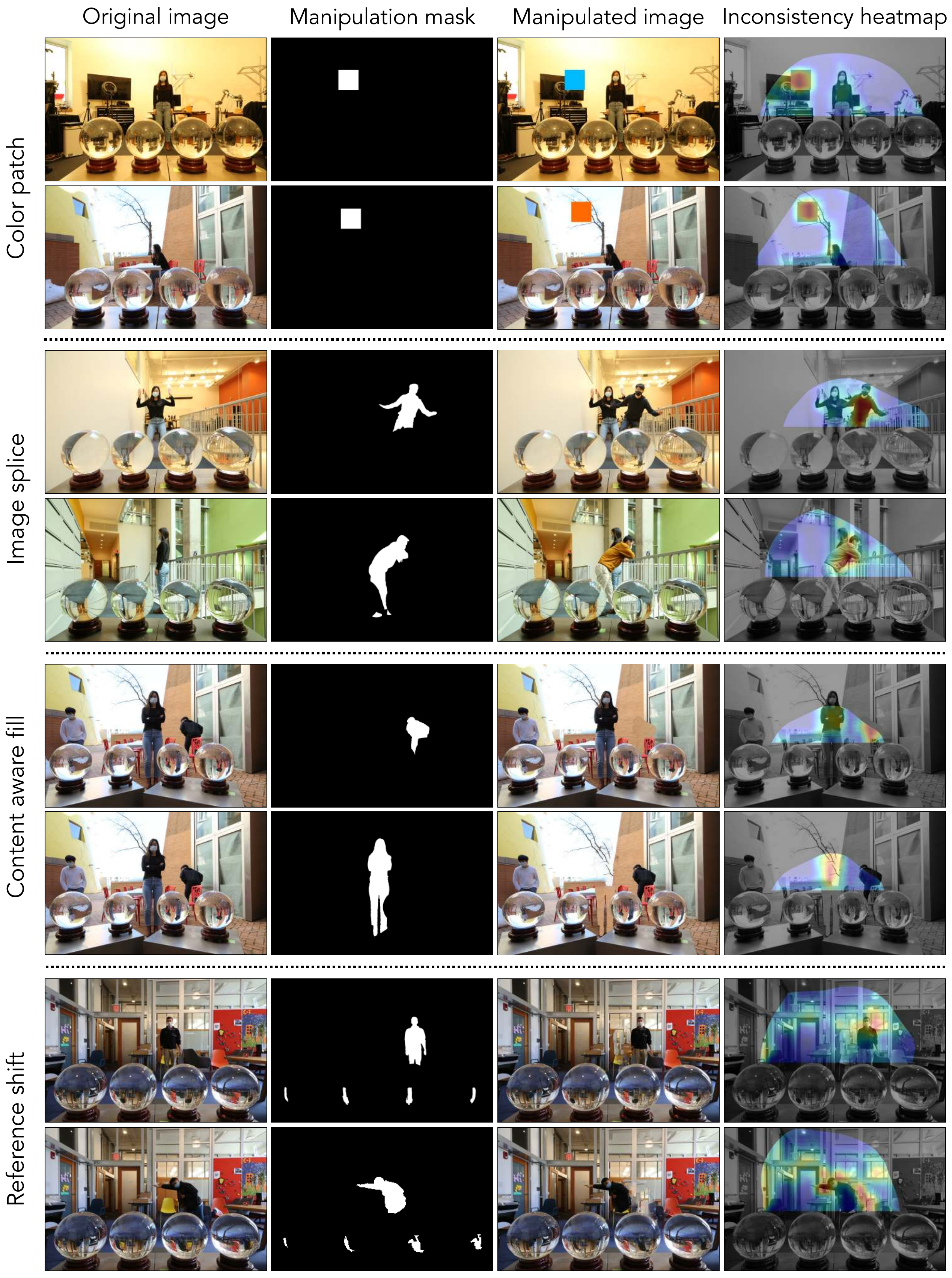}
\vspace{-7pt}\caption{{\textbf{Additional results.} Additional detection results for different scenes and different types of manipulations. Our method compares the totem protected region of the scene reconstruction with the manipulated image and shows potential manipulations via the inconsistency heatmap. Note that the scene reconstruction is learned only using the pixels within the totems.}}
\label{fig:supp-results-more}
\vspace{-7pt}
\end{figure}

\subsection{Totem configuration} 
Totems can be placed anywhere between the camera and the scene region to be protected (\eg the subject). We show reconstruction and detection results (Fig.~\ref{fig:supp-results}) for the same scene and manipulation type (CAF) while varying the totem configuration. 

Note that configurations that contain totems farther from the camera (row 1 and 4) have smaller protected regions. This is not due to reduced reconstruction quality (column 3), but because we use the same density threshold (Sec \ref{manip-detect} main text) for images with less number of totem pixels (\ie overall lower projection density). Future work can explore a less rigorous threshold strategy that takes the total amount of totem pixels into account.

\subsection{Additional results}
In Fig.~\ref{fig:supp-results-more}, we show additional detection results for the following manipulations: 1) inserting randomly colored patches, 2) adding people by image splicing, 3) removing people with Photoshop CAF, or 4) shifting people in both camera and totem views to the same reference position. Our method measures the patch-wise L1 distance between the protected region of the reconstruction and the manipulated image and shows a heatmap that highlights potential manipulations.

\section{Scope of the totem framework}\label{sec:sm_scope}

The goal of the totem framework is to propose a novel geometric and physical approach to image forensics, demonstrate its potential, and inspire further cross-domain research. While our current method and choice of totems cannot yet defend all types of manipulations, we discuss specific manipulation settings and use cases below.

\subsection{Discussion of possible attacks}
\myparagraph{Image manipulation.} Totem identities are unknown to the adversary a priori. Spherical totems are more visible due to their interpretable distortion patterns. This prompts the adversary to manipulate the totem views to avoid being detected under human inspection. In an ideal scenario, a totem with more complex and compact geometry would be less noticeable and ultimately less likely to be manipulated by the adversary. For such a case, we have demonstrated through many examples that when only the image is manipulated and totem views remain intact, our method can reliably detect a variety of manipulations (\ie color patches, image splice, CAF).

\myparagraph{Joint image and totem manipulations.} If the adversary notices and attempts to manipulate the totem views, there are a few different possibilities:

\begin{itemize}%
  \item \textbf{Cropping out totems.} In this case, the image is no longer verifiable; only verifiable images are protected.
  \item \textbf{Scrambling totem pixels.} Our method can still reliably reconstruct the scene when small portions of the totem views are manipulated (\eg the reference shift examples). If the resulted reconstruction seems drastically different from the camera view, it implies that large portions of the totem views have been manipulated. 
  \item \textbf{Geometric manipulation.} We demonstrate that geometric manipulation of the totem views is detectable through the reference shift example in Fig.\ref{fig:supp-results-more} and Fig.\ref{fig:detection} in the main paper. The adversary shifts the subject in both camera and totem views to the same reference position in the scene. The resultant manipulation seems reasonable under human inspection but creates geometric inconsistency and makes the reconstructed subject distorted. The reconstruction disagrees with the manipulated camera view, making this manipulation detectable.
  \item \textbf{Color manipulation.} If the adversary changes the color of an object (\ie jacket) in both camera and totem views without tampering with the geometry, the reconstruction will contain the manipulated color and agree with the manipulated camera view. This is a case where our method can fail.
\end{itemize}

\myparagraph{Limitations.} Currently, our method reliably detects manipulations of big objects (\ie the entire subject). As research in sparse-view scene reconstruction continues to develop ~\cite{Niemeyer2021Regnerf},  we expect improved reconstruction results with less noise and more semantic details, allowing more detailed manipulations such as smaller objects or facial expressions to be detected. Another limitation is that our detection method is designed to highlight discrepancy between the reconstruction and the camera view, which means it currently does not highlight manipulations in the totem views. This is important to address in future work.

\subsection{Frequently asked questions}

\myparagraph{Who owns/uses totems?} The subject owns and sets up their unique totems as a manipulation defense for the digital content captured by anyone or any device. A common confusion is that the photographer carries totems around and is responsible for setting up totems. While there are many active defense methods available to other stakeholders (\eg digital signatures, encryption cameras, etc.), method designed for the subject is under-explored and we hope to inspire more research in this area.

\myparagraph{Why not treat totems as cameras and use Structure from Motion (SfM)?} SfM and other methods that rely on correspondences will not generalize to totems with complex geometry and distortions.

\end{document}